\documentclass{article}
\usepackage{spconf,amsmath,graphicx}
\usepackage{times}
\usepackage{epsfig}
\usepackage{amssymb}
\usepackage{booktabs}
\usepackage{multirow}
\usepackage{subcaption}
\usepackage[export]{adjustbox}
\usepackage{float}
\usepackage{dashrule}
\usepackage{arydshln}
\usepackage[utf8x]{inputenc}

\title{Adversarial Segmentation Loss for Sketch Colorization}

\name{Samet Hicsonmez$^{\star}$ \qquad Nermin Samet$^{\dagger}$ \qquad Emre Akbas$^{\dagger}$ \qquad Pinar Duygulu$^{\star}$}
  \address{$^{\star}$ Computer Engineering, Hacettepe University, Ankara, Turkey \\
     $^{\dagger}$ Computer Engineering, Middle East Technical University, Ankara, Turkey}

\begin{document}
\ninept

\maketitle

\begin{abstract}
We introduce a new method for generating color images from sketches or edge maps. Current methods either require some form of additional user-guidance or are limited to the ``paired’’ translation approach. We argue that segmentation information could provide valuable guidance for sketch colorization. To this end, we propose to leverage semantic image segmentation, as provided by a general purpose panoptic segmentation network, to create an additional adversarial loss function. Our loss function can be integrated to any baseline GAN model. Our method is not limited to datasets that contain segmentation labels, and it can be trained for ``unpaired’’ translation tasks. We show the effectiveness of our method on four different datasets spanning scene level indoor, outdoor, and children book illustration images using qualitative, quantitative and user study analysis. Our model improves its baseline up to 35 points on the FID metric. Our code and pretrained models can be found at https://github.com/giddyyupp/AdvSegLoss.

\end{abstract}

\begin{keywords}
sketch colorization, sketch to image translation, Generative Adversarial Networks (GAN), image segmentation, image to image translation

\end{keywords}

\section{Introduction}
\label{sec:intro}

Generating an image from an input sketch (i.e. edge map), a task known as ``sketch to image translation’’, or ``sketch colorization’’ for short, is an attractive task as sketches are easy to obtain and convey essential information about the content of an image. At the same time, it is a challenging task due to the large domain gap between single channel edge maps and color images. In addition, sketches usually lack details for background objects, and even sometimes for foreground objects.

Sketch colorization has been explored in a variety of domains including faces~\cite{deepfacedrawing,lee2020reference,linestofacephoto}, objects~\cite{sketchygan,texturegan,lu2018image,liuunsupervised}, animes~\cite{comicolorization,ci2018user,autopainter,zhang2018two,zhang2017style}, art~\cite{liu2020sketch} and scenes~\cite{scribbler,edgegan,zou2019language}. Most of the methods in these studies require some form of user-guidance, as additional input, in the form of, e.g., a reference color, patch or image. Without this guidance, these methods produce unrealistic colorizations. Another important observation is that, except Liu et al.’s work~\cite{liuunsupervised}, all methods follow the ``paired’’ approach, which limits the method to datasets that have a ground-truth image per sketch.

In this work, we propose to leverage general purpose semantic image segmentation to alleviate the two shortcomings mentioned above. Semantic segmentation methods have matured to a level that they produce useful results even for datasets on which they were not trained (Section~\ref{sec:dataset}). We hypothesize that a correctly colored sketch would yield a proper segmentation result and leverage this result in an extra adversarial loss in a GAN setting. By doing so, our method neither requires additional user guidance nor becomes limited to the ``paired’’ domain.

We propose a new method which utilizes semantic segmentation for sketch based image colorization problem. We introduce three models considering different levels of segmentation feedback in the sketch to image translation pipeline. Our models could be integrated into any paired or unpaired GAN model.
We demonstrate effectiveness of using segmentation cues through extensive evaluations. Our contributions in this paper can be summarized as follows. (i) We propose to use general purpose semantic segmentation as an additional adversarial loss in a GAN model, for the sketch colorization problem. Ground-truth segmentation labels are not a requirement for our approach. (ii) Our method is neither specific to a domain (e.g. face, art, anime, etc.) nor limited to the ``paired’’ approach. (iii) We conduct extensive evaluations on four distinct datasets on both paired and unpaired settings, and show the effectiveness of our method through qualitative, quantitative and user study analysis.

\begin{figure*}[h]
\centering
\includegraphics[width=0.75\textwidth]{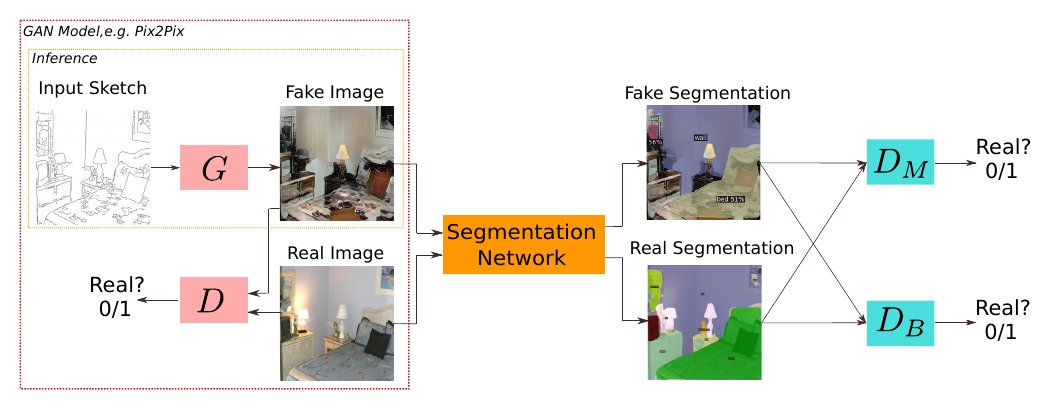}
\caption{Our proposed model with adversarial segmentation loss for sketch colorization.}
\label{fig:model}
\end{figure*}
\section{Related Work}
\label{sec:related_work}

Even though the edge map and sketch of an image are different concepts, in practice, XDoG~\cite{xdog} or HED~\cite{hed} based edge maps are considered as sketches (e.g.,~\cite{scribbler, autopainter}). Moreover, some sketch based models~\cite{sketchygan} use edge maps for data augmentation. Hence, we refer to all these methods as sketch to image translation models. General purpose image-to-image translation methods~\cite{cyclegan, dualgan, pix2pix, ganilla, huang2018multimodal} could be used to solve sketch to image translation tasks. However, the results of these generic methods are usually not very satisfactory.

One widely used solution to improve the colorization performance is to employ additional  color~\cite{scribbler,autopainter,zhang2018two,ci2018user}, patch~\cite{texturegan}, image~\cite{lee2020reference,zhang2017style,liu2020sketch,comicolorization} or language guidance~\cite{zou2019language}. For instance, in color guidance, users specify their desired colors for the regions in the sketch image, and the model utilizes this information to generate exact or similar colors for these regions. Some automatic methods also utilize user guidance to improve their performance resulting in a hybrid approach. Most of the sketch to image translation methods are based on the ``paired’’ training approach~\cite{scribbler, autopainter, sketchygan, edgegan}, however, recently unpaired methods have also been presented~\cite{liuunsupervised, liu2020sketch}.

In Scribbler~\cite{scribbler}, one of the very first paired and user guided scene sketch colorization models, in addition to pixel, perceptual and GAN losses, they use total variation loss to encourage smoothness. They use XDoG to generate sketch images of 200k bedroom photos, and produce $128\times128$ colorized images.
Zou et al.~\cite{zou2019language} use text inputs to progressively colorize an input sketch, in such a way that a novel text guided sketch segmenter segments and locates the objects in the scene.
EdgeGAN~\cite{edgegan} maps edge images to a latent space during training using an edge encoder. During inference, the edge encoder encodes the input sketch to the latent space to subsequently generate a color image. They experimented with 14 foreground and 3 background objects from COCO~\cite{coco} dataset.

EdgeGAN~\cite{edgegan} and Scribbler~\cite{scribbler} use a supervised approach where input sketches and corresponding output images exist. However, it is hard to collect sketch image pairs.
Liu et al.~\cite{liuunsupervised} propose a two stage method to convert object sketches to color images in an unsupervised (unpaired) way. They first convert sketches to grayscale images, then color images. Self supervision is also used to complete the deliberately deleted sketch parts and clear the added noisy edges from sketch images. In Sketch-to-Art~\cite{liu2020sketch}, an art image is generated using an input sketch and a target art style image. They encode content of the input sketch and style of the art image, then fuse both features to generate a stylized art image.

\section{Model}
\label{sec:model}

Figure~\ref{fig:model} shows the overall structure of our proposed model. The box with dashed yellow borders shows the inference stage of our model. Red border marks the GAN model used for sketch to image translation. In this work, we used \textit{Pix2Pix} and \textit{CycleGAN} as baselines for paired and unpaired training, respectively. This preference is made based on the effectiveness of both methods across a variety of tasks and datasets. Our model could be integrated into any other GAN model.

Our model consists of a baseline GAN, a panoptic segmentation network (\textit{Seg}) and two discriminators (\textit{$D_{M}$} and \textit{$D_{B}$}). Panoptic segmentation network is trained offline on the COCO Stuff~\cite{coco_stuff} dataset and 
its weights are frozen during the training of our model. Real and fake images are fed to the \textit{Seg} network to get real and fake segmentation maps. Then, these two segmentation maps are given to the discriminators to classify them as real or fake.

We designed three variants of our model to embed different levels of segmentation feedback to the sketch to image translation pipeline.

The first variant utilizes the full segmentation map of an image where all foreground and background classes -- a total of 135 classes -- are considered. In this model, ground-truth color image $I_{real}$ and the generated color image $I_{fake}$ are fed to \textit{Seg} which outputs full segmentation maps for both images. Then, these two outputs are given to a discriminator network \textit{$D_{M}$} to discriminate between real and fake segmentation maps. We call this model as \textit{Multi-class} in the rest of the paper.

As a higher level of abstraction, grouping objects only as background and foreground may yield sufficient information. In the second variant of our model, we only used two classes (background and foreground) in the segmentation map by grouping all foreground classes into one and all background classes into another class. As with our original multi-class model, binary segmentation outputs for real and fake images are fed to a discriminator network \textit{$D_{B}$} to discriminate between real and fake ones. We refer to this model as \textit{Binary}.

Finally, our third variant is the union of the above two. It contains both discriminators, and is named as \textit{Both}. Overall loss function for our model is the summation of losses of the baseline GAN model ($L_G$) and the two additional discriminators’ ($L_B$ and $L_M$). Formally, the objective function is $\mathcal{L} = w_gL_G + w_bL_B + w_mL_M$. We run ablation experiments to set the best values of $w_g$, $w_b$ and $w_m$. We keep $w_g$ fixed and search for $w_b$ and $w_m$ on the bedroom dataset using the \textit{Both} model setting. Results show that using $1.0$ for $w_g$, $w_b$ and $w_m$ yields the best FID score (see Table~\ref{tab:abl}).

We used PyTorch~\cite{pytorch} to implement our models.
We use sketch images as source domain, and color images as target domain. Datasets are described in Section~\ref{sec:dataset}. All training images (i.e. color and sketch images) are resized to $256\times256$ pixels. We train all models for $200$ epochs using the Adam optimizer~\cite{adam} with a learning rate of $0.0002$. We conducted all our experiments on a Nvidia Tesla V100 GPU.

\begin{figure}
\captionsetup[subfigure]{labelformat=empty}
\centering
\begin{subfigure}[b]{0.10\textwidth}
\includegraphics[width=1.0\textwidth]{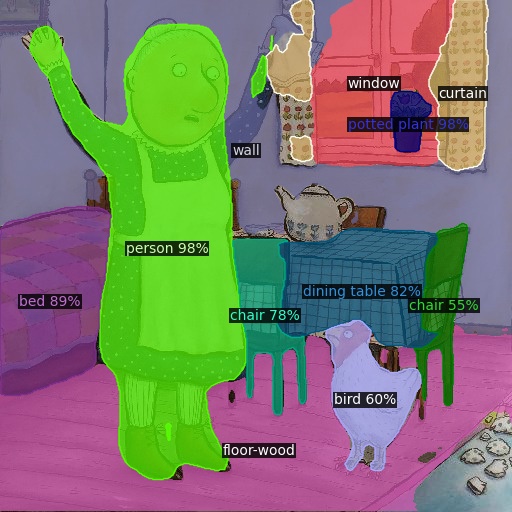}
 \vspace{0.05 cm}
\includegraphics[width=1.0\textwidth]{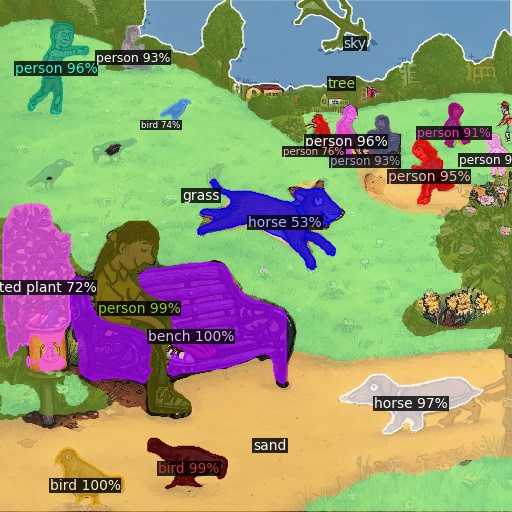}
\end{subfigure}
\begin{subfigure}[b]{0.10\textwidth}
\includegraphics[width=1.0\textwidth]{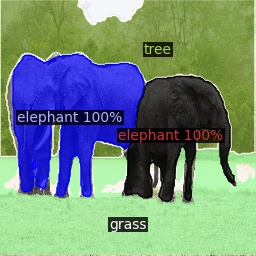}
 \vspace{0.05 cm}
\includegraphics[width=1.0\textwidth]{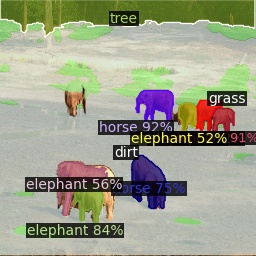}
\end{subfigure}
\begin{subfigure}[b]{0.10\textwidth}
\includegraphics[width=1.0\textwidth]{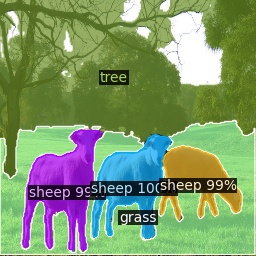}
 \vspace{0.05 cm}
\includegraphics[width=1.0\textwidth]{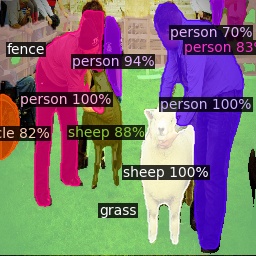}
\end{subfigure}
\begin{subfigure}[b]{0.15\textwidth}
\includegraphics[width=1.0\textwidth]{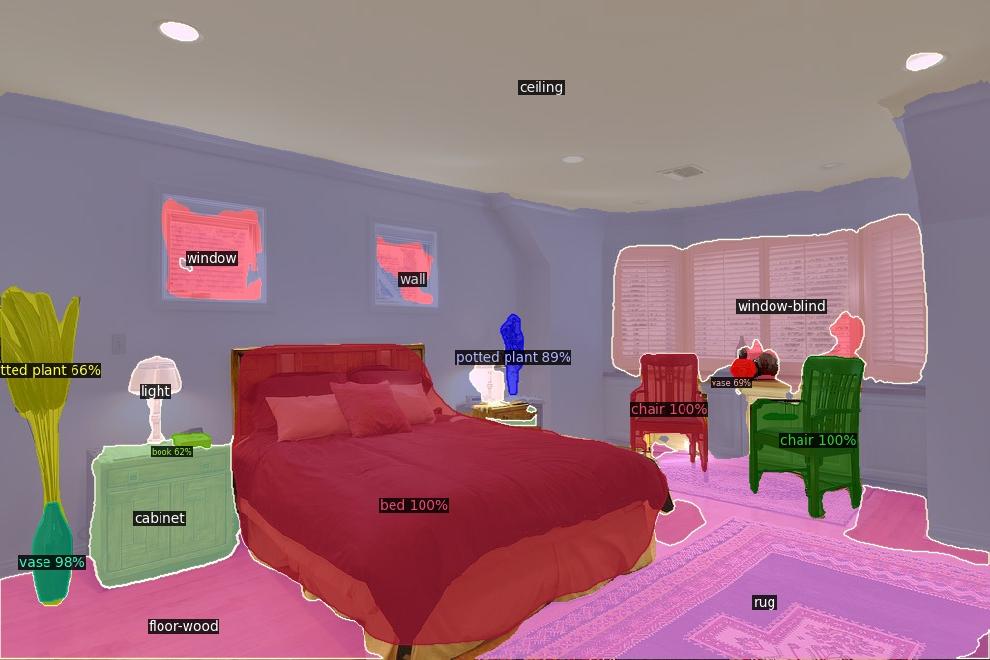}
\vspace{0.05 cm}
\includegraphics[width=1.0\textwidth]{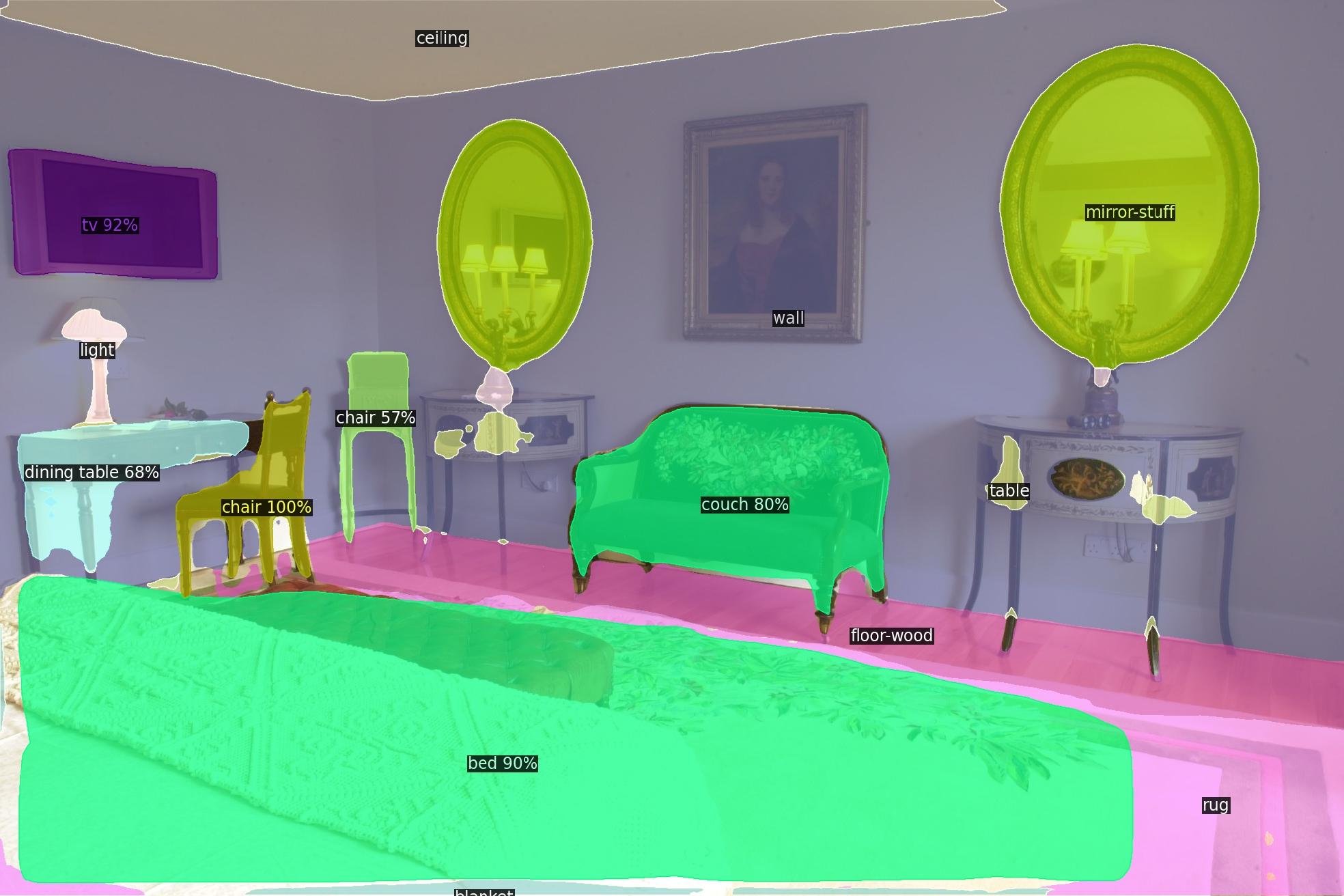}
\end{subfigure}
\caption{Sample segmentation results on different datasets.}
\label{fig:seg-samples}
\end{figure}

\begin{figure*}[ht]
\centering
\begin{tabular}{cccccc}
\shortstack{Input \\ { }} & \shortstack{GT \\ { }} & \shortstack{Baseline \\ { }}  & \shortstack{AdvSegLoss \\ (Multi-class)} & \shortstack{AdvSegLoss\\ (Binary)} & \shortstack{AdvSegLoss\\(Both)} \\
\includegraphics[width=0.13\textwidth,  ,valign=m, keepaspectratio,]{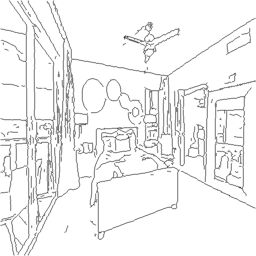}  \hspace*{-12pt} &
\includegraphics[width=0.13\textwidth,  ,valign=m, keepaspectratio,]{./figures/visual_res/ade20k/unpaired/gt/ADE_val_00000129.jpg} \hspace*{-12pt} &
\includegraphics[width=0.13\textwidth,  ,valign=m, keepaspectratio,]{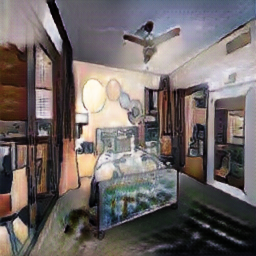} \hspace*{-12pt} &
\includegraphics[width=0.13\textwidth,   ,valign=m, keepaspectratio,]{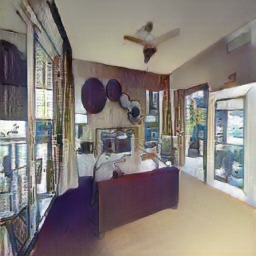} \hspace*{-12pt} &
\includegraphics[width=0.13\textwidth,   ,valign=m, keepaspectratio,]{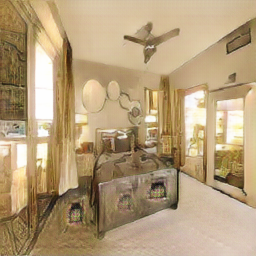} \hspace*{-12pt} &
\includegraphics[width=0.13\textwidth,   ,valign=m, keepaspectratio,]{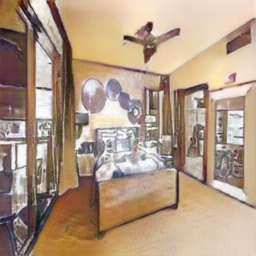} \hspace*{-12pt} \\
\noalign{\vskip 2mm}

\includegraphics[width=0.13\textwidth,  ,valign=m, keepaspectratio,]{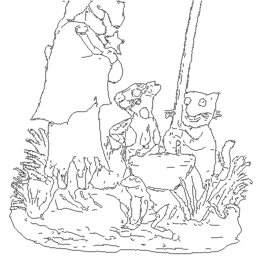}  \hspace*{-12pt}&
\includegraphics[width=0.13\textwidth,  ,valign=m, keepaspectratio,]{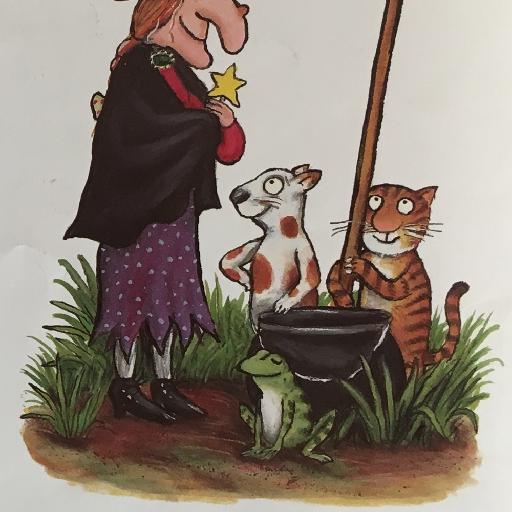} \hspace*{-12pt}&
\includegraphics[width=0.13\textwidth,  ,valign=m, keepaspectratio,]{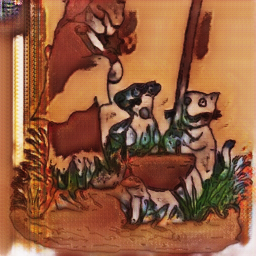} \hspace*{-12pt}&
\includegraphics[width=0.13\textwidth,   ,valign=m, keepaspectratio,]{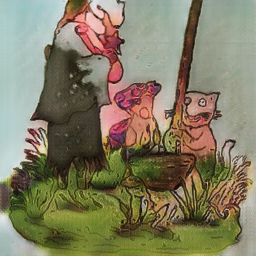} \hspace*{-12pt} &
\includegraphics[width=0.13\textwidth,   ,valign=m, keepaspectratio,]{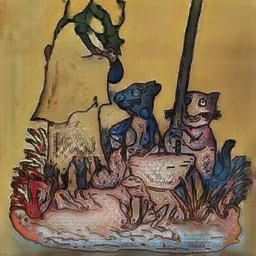} \hspace*{-12pt} &
\includegraphics[width=0.13\textwidth,   ,valign=m, keepaspectratio,]{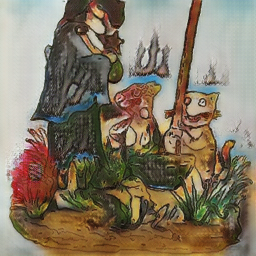} \hspace*{-12pt}\\

\midrule
\includegraphics[width=0.13\textwidth,  ,valign=m, keepaspectratio,]{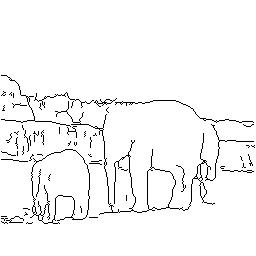}  \hspace*{-12pt}&
\includegraphics[width=0.13\textwidth,  ,valign=m, keepaspectratio,]{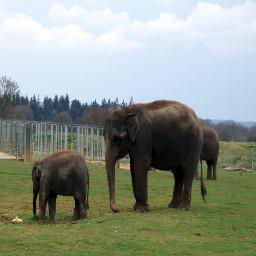} \hspace*{-12pt}&
\includegraphics[width=0.13\textwidth,  ,valign=m, keepaspectratio,]{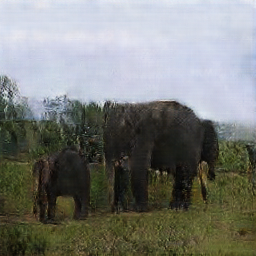} \hspace*{-12pt}&
\includegraphics[width=0.13\textwidth,   ,valign=m, keepaspectratio,]{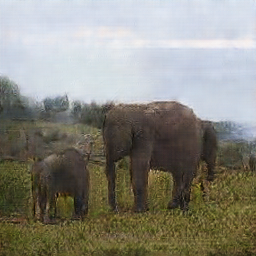} \hspace*{-12pt} &
\includegraphics[width=0.13\textwidth,   ,valign=m, keepaspectratio,]{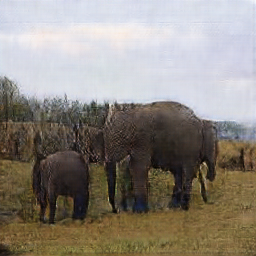} \hspace*{-12pt} &
\includegraphics[width=0.13\textwidth,   ,valign=m, keepaspectratio,]{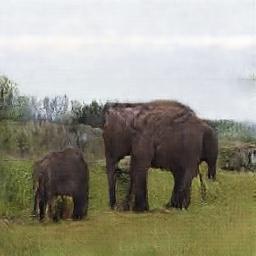} \hspace*{-12pt}\\
\noalign{\vskip 2mm}

\includegraphics[width=0.13\textwidth,  ,valign=m, keepaspectratio,]{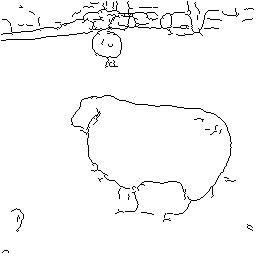}  \hspace*{-12pt}&
\includegraphics[width=0.13\textwidth,  ,valign=m, keepaspectratio,]{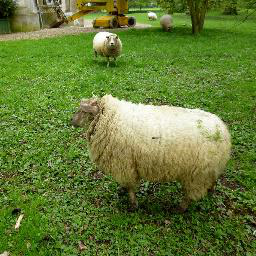} \hspace*{-12pt}&
\includegraphics[width=0.13\textwidth,  ,valign=m, keepaspectratio,]{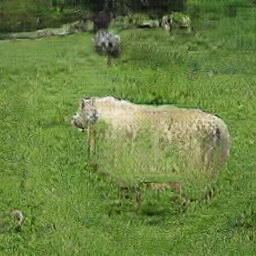} \hspace*{-12pt}&
\includegraphics[width=0.13\textwidth,   ,valign=m, keepaspectratio,]{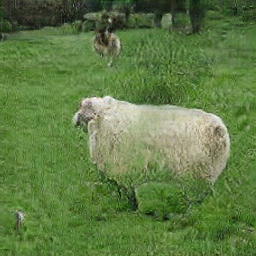} \hspace*{-12pt} &
\includegraphics[width=0.13\textwidth,   ,valign=m, keepaspectratio,]{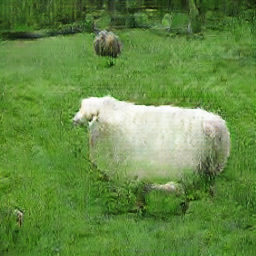} \hspace*{-12pt} &
\includegraphics[width=0.13\textwidth,   ,valign=m, keepaspectratio,]{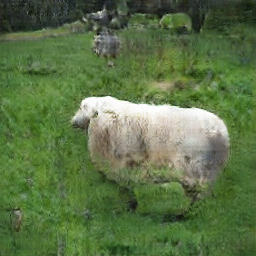} \hspace*{-12pt}  \\
\end{tabular}
\caption{Sample results from baselines (CycleGAN and Pix2Pix) and our model with different settings. Input images on each row are from bedroom, illustration, elephants and sheep datasets, respectively. First two rows display results of unpaired training, and last two rows show results for paired training.
On bedroom and elephant datasets \textit{Binary}, on illustration and sheep datasets \textit{Both} setting gave best results for both training schemes.}
\label{fig:all-results}
\end{figure*}

\section{Dataset}
\label{sec:dataset}

We evaluated our models on four challenging datasets. The first dataset consists of bedroom images from the Ade20k indoor dataset \cite{ade20k}, with $1355$ train and $135$ test images. The second dataset contains children’s book illustrations by Alex Scheffler~\cite{ganilla}, with $659$ train and $131$ test images. The third and fourth datasets were curated by us from the COCO dataset. We collected images which contain elephant or sheep instances. Note that these images contain not only elephants and sheeps but objects/regions from other foreground and background classes such as person, animals, mountains, grass and sky. Elephant dataset contains $1800$ train and $343$ test images, and the sheep dataset has $1300$ train and $229$ test images.
Example images from these datasets and their segmentation outputs are shown in Figure~\ref{fig:seg-samples}.

Edge images are extracted using the HED~\cite{hed} method. In the first two columns of Figure~\ref{fig:all-results}, we present sample natural and edge images for all the datasets. It can be seen that the images contain a variety of foreground and background objects, also it is hard -- even for the trained eye -- to figure out the source dataset for some of the edge images.
Our code, pretrained models and the scripts to produce the ``sheep’’ and ``elephant’’ datasets, and corresponding sketch images can be found at https://github.com/giddyyupp/AdvSegLoss.

\section{Experiments}
\label{sec:exp}

We compared our models with the baseline models: CycleGAN~\cite{cyclegan}, AutoPainter~\cite{autopainter} and Pix2Pix~\cite{pix2pix}. We used their official implementations that are publicly available. Baseline models are trained for $200$ epochs.

\begin{table*}
\begin{center}
\caption{Comparison with baseline methods in terms of FID scores, lower is better.} 
\label{tab:fid_scores}
\resizebox{\textwidth}{!}{
\begin{tabular}{lcccc:ccccc}
 \toprule 
 \multirow{3}{*}{Dataset} &  \multicolumn{4}{c}{Unpaired} &  \multicolumn{5}{c}{Paired} \\
\cmidrule(lr){2-5}  \cmidrule(lr){6-10}
& \shortstack{CycleGAN\\ { }} & \shortstack{+AdvSegLoss\\(Multi-class)} & \shortstack{+AdvSegLoss\\(Binary)} & \shortstack{+AdvSegLoss\\(Both)} & \shortstack{AutoPainter \\ { }} & \shortstack{Pix2Pix \\ { }} & \shortstack{+AdvSegLoss\\(Multi-class)} & \shortstack{+AdvSegLoss\\(Binary)} & \shortstack{+AdvSegLoss\\(Both)}\\
 \midrule 
Bedroom         & 113.1 & 111.7 & \textbf{87.1} & 93.2           & 206.8 & 100.5 & 100.0 & \textbf{95.1} & 110.1  \\
Illustration         & 213.6 & 206.9 & 204.8 & \textbf{189.4}           & 272.0 &180.0 & 176.9 & 178.0 & \textbf{175.7}   \\
Elephant        & 126.4 & 103.9 & \textbf{91.9} & 116.9        & 155.1 & 83.5   & 85.8   & \textbf{78.8} & 82.8      \\
Sheep                & 209.3 & 207.2 & 236.1 & \textbf{196.8}          & 233.1 & 157.0 & 159.9 & 162.0 & \textbf{150.5}   \\
\bottomrule 
\end{tabular}}
\end{center}
\end{table*}

\subsection{Quantitative Analysis}

To quantitatively evaluate the quality of generated images, we used the widely adopted Frechet Inception Distance (FID)~\cite{FID} metric. FID score measures the distance between the distributions of the generated and real images.  Lower FID score indicates the higher similarity between two image sets.
We present FID scores for all the experiments in the Table~\ref{tab:fid_scores}. FID scores are inline with the visual inspections (see Figure~\ref{fig:all-results}), for all the datasets, at least one of the variants of our model performed better than the baseline.

First of all, when we compare FID scores of two training schemes and baseline models, paired training (Pix2Pix) performed better than unpaired training, as expected. However, our ``adversarial segmentation loss’’ affected the results of paired and unpaired cases differently. For instance, on elephant dataset our models improved baseline up to $35$ points for unpaired case, but only $5$ points for paired case.

Another crucial observation is that segmentation guidance closed the gap between unpaired and paired training results. Best FID scores for unpaired models on bedroom, illustration and elephant datasets become very close to or even better than paired training. For instance on the elephant dataset, the initial $40+$ point FID gap ($126$ vs $83$) dropped to $13$ ($92$ vs $79$) on \textit{Binary} setting. Here the only exception is the sheep dataset. Since the sheep dataset contains various complex objects, unpaired and paired models failed to generate plausible images.

When we look at the best performing settings on different datasets, structure of the dataset has an effect on the results. For instance, even though one is an indoor and the other one is an outdoor dataset, bedroom and elephant images are composed of similar structure. FG/BG ratios and placements of them in these datasets are similar across all images, i.e. walls, ceiling and floors in bedroom images are always positioned in the same places on different images. Also elephant images contain very few FG objects, i.e. only elephants most of the time, and large BG areas such as grass, trees and sky. On these two datasets, \textit{Binary} setting which considers FG/BG classes only gave the best FID score.
On the other hand, illustration and sheep images got a variety of FG objects and scenes. On such datasets, using only a FG/BG discriminator even degrades the performance.

\begin{table}[h]
\begin{minipage}[c]{0.33\columnwidth}
\centering
\captionof{table}{\\Ablation results.}
\label{tab:abl}
\resizebox{0.99\columnwidth}{!}{
\begin{tabular}{lcc}
\toprule 
 $w_b$ and $w_m$   & FID \\
 \midrule 
0.1 & 114.8  \\
0.5 & 114.5  \\
1.0 & \textbf{93.2}  \\
5.0 & 147.8  \\
10.0 & 104.6  \\
\bottomrule 
\end{tabular} } 
\end{minipage}
\begin{minipage}[l]{0.60\columnwidth}
\centering
\captionof{table}{User Study results.}
\label{tab:user_study}
\resizebox{0.99\columnwidth}{!}{
\begin{tabular}{lcc}
\toprule 
 Dataset & CycleGAN & +AdvSegLoss \\
 \midrule 
Bedroom         & 20.0 & \textbf{80.0}  \\
Illustration        & 27.0 & \textbf{73.0}  \\
Elephant        & 39.1 & \textbf{60.9}  \\
Sheep                & 19.1 & \textbf{80.9}  \\
 \bottomrule 
\end{tabular} } 
\end{minipage}
\end{table}

\subsection{Qualitative Analysis and User Study}

We present visual results of sketch colorization for our model and the baseline models in the Figure~\ref{fig:all-results}.
On bedroom and illustration datasets, we show results of unpaired training, and on elephant and sheep datasets we show paired training results.

On the bedroom dataset, the \textit{Binary} setting generates better images compared to baselines and other settings. Colors are uniform across the object parts in this setting. There are defective colors in the CycleGAN results such as the bottom of the bed and floor.
On the illustration dataset, the baseline model performed poorly. Objects are hard to recognize and most importantly colors are not proper at all. On the other hand, \textit{Multi-class} and \textit{Both} settings generate significantly better images i.e. generated objects and background got consistent colors.

Finally, on elephant and sheep datasets, although generated images are not very visually appealing for all the methods, segmentation guided images are quite appealing compared to baseline models’. On the elephant dataset \textit{Binary}, on the sheep dataset \textit{Both} setting performed the best.

We conducted a user study\footnote{Readers could reach our study at http://52.186.136.234:8080/} to measure realism of generated images. We show two random images (at random positions, left or right) which were generated with CycleGAN and our best setting (lowest FID score) for all four datasets, and asked participants to select the more realistic one.

We collected a total of 115 survey inputs from 39 different users. In Table~\ref{tab:user_study}, we present results of the user study in terms of preference percentages of each model. User study results are inline with the FID score results, on all datasets, images generated by our model were preferred by the users most of the time.

\begin{figure}
\captionsetup[subfigure]{labelformat=empty}
\centering
\begin{subfigure}[b]{0.10\textwidth}
\caption{Input}
\includegraphics[width=1.0\textwidth]{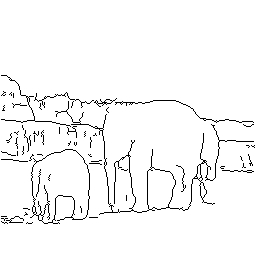}
\vspace{0.05 cm}
\includegraphics[width=1.0\textwidth]{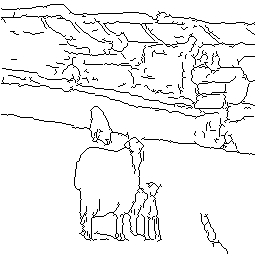}
\vspace{0.05 cm}
\end{subfigure}
\begin{subfigure}[b]{0.10\textwidth}
\caption{GT}
\includegraphics[width=1.0\textwidth]{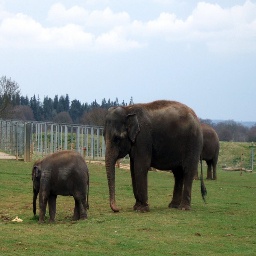}
\vspace{0.05 cm}
\includegraphics[width=1.0\textwidth]{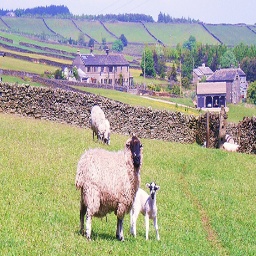}
\vspace{0.05 cm}
\end{subfigure}
\begin{subfigure}[b]{0.10\textwidth}
\caption{CycleGAN}
\includegraphics[width=1.0\textwidth]{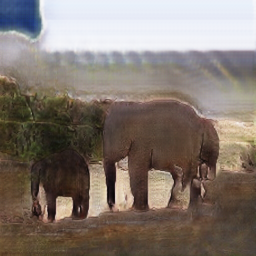}
\vspace{0.05 cm}
\includegraphics[width=1.0\textwidth]{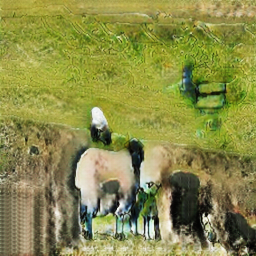}
\vspace{0.05 cm}
\end{subfigure}
\begin{subfigure}[b]{0.10\textwidth}
\caption{+AdvSegLoss}
\includegraphics[width=1.0\textwidth]{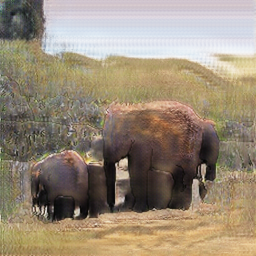}
\vspace{0.05 cm}
\includegraphics[width=1.0\textwidth]{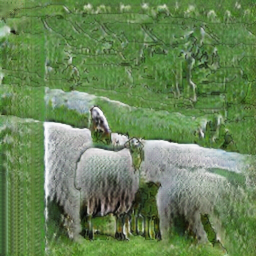}
\vspace{0.05 cm}
\end{subfigure}
\caption{Sample results on elephant and sheep datasets for CycleGAN and our best model. Realism of both models are not satisfactory, however, especially colors of BG areas are better in our results. }
\label{fig:realism-res}
\end{figure}

\section{Conclusion}
\label{sec:conclusion}

In this paper, we presented a new method for the sketch colorization problem. Our method utilizes a general purpose image segmentation network and adds an adversarial segmentation loss (AdvSegLoss) to the regular GAN loss. AdvSegLoss could be integrated to any GAN model, and works even if the dataset doesn’t have segmentation labels.  We used CycleGAN and Pix2Pix as baseline GAN models in this work.
We conduct extensive evaluations on various datasets including bedroom, sheep, elephant and illustration images and evaluate the performance both quantitatively (using FID score) and qualitatively (through a user study). We show that our model outperforms baselines on all datasets on both FID score and user study analysis.

Regarding the limitations of our method, although we improve the baseline both qualitatively and quantitatively, especially elephant and sheep results lack realism. Even the paired training results are not visually appealing on these two datasets (last two rows of Figure~\ref{fig:all-results}), most probably due to the fact that the baseline models are not very successful at generating complex scenes.

\section{Acknowledgements}
The numerical calculations reported in this paper were fully performed at TUBITAK ULAKBIM, High Performance and Grid Computing Center (TRUBA resources).

\ninept

\bibliographystyle{IEEEbib}
\bibliography{refs}

\begin{thebibliography}{10}

\bibitem{deepfacedrawing}
Shu-Yu Chen, Wanchao Su, Lin Gao, Shihong Xia, and Hongbo Fu,
\newblock ``Deepfacedrawing: deep generation of face images from sketches,''
\newblock {\em ACM Transactions on Graphics (TOG)}, vol. 39, no. 4, pp. 72--1,
  2020.

\bibitem{lee2020reference}
Junsoo Lee, Eungyeup Kim, Yunsung Lee, Dongjun Kim, Jaehyuk Chang, and Jaegul
  Choo,
\newblock ``Reference-based sketch image colorization using augmented-self
  reference and dense semantic correspondence,''
\newblock in {\em CVPR}, 2020.

\bibitem{linestofacephoto}
Yuhang Li, Xuejin Chen, Feng Wu, and Zheng-Jun Zha,
\newblock ``Linestofacephoto: Face photo generation from lines with conditional
  self-attention generative adversarial networks,''
\newblock in {\em ACM MM}, 2019.

\bibitem{sketchygan}
Wengling Chen and James Hays,
\newblock ``Sketchygan: Towards diverse and realistic sketch to image
  synthesis,''
\newblock in {\em CVPR}, 2018.

\bibitem{texturegan}
Wenqi Xian, Patsorn Sangkloy, Varun Agrawal, Amit Raj, Jingwan Lu, Chen Fang,
  Fisher Yu, and James Hays,
\newblock ``Texturegan: Controlling deep image synthesis with texture
  patches,''
\newblock in {\em CVPR}, 2018.

\bibitem{lu2018image}
Yongyi Lu, Shangzhe Wu, Yu-Wing Tai, and Chi-Keung Tang,
\newblock ``Image generation from sketch constraint using contextual gan,''
\newblock in {\em ECCV}, 2018.

\bibitem{liuunsupervised}
Runtao Liu, Qian Yu21, and Stella~X Yu,
\newblock ``Unsupervised sketch to photo synthesis,''
\newblock in {\em ECCV}, 2020.

\bibitem{comicolorization}
Chie Furusawa, Kazuyuki Hiroshiba, Keisuke Ogaki, and Yuri Odagiri,
\newblock ``Comicolorization: semi-automatic manga colorization,''
\newblock in {\em SIGGRAPH Asia 2017 Technical Briefs}. 2017.

\bibitem{ci2018user}
Yuanzheng Ci, Xinzhu Ma, Zhihui Wang, Haojie Li, and Zhongxuan Luo,
\newblock ``User-guided deep anime line art colorization with conditional
  adversarial networks,''
\newblock in {\em ACM MM}, 2018.

\bibitem{autopainter}
Yifan Liu, Zengchang Qin, Tao Wan, and Zhenbo Luo,
\newblock ``Auto-painter: Cartoon image generation from sketch by using
  conditional wasserstein generative adversarial networks,''
\newblock {\em Neurocomputing}, vol. 311, pp. 78 -- 87, 2018.

\bibitem{zhang2018two}
Lvmin Zhang, Chengze Li, Tien-Tsin Wong, Yi~Ji, and Chunping Liu,
\newblock ``Two-stage sketch colorization,''
\newblock {\em ACM Transactions on Graphics (TOG)}, vol. 37, no. 6, pp. 1--14,
  2018.

\bibitem{zhang2017style}
Lvmin Zhang, Yi~Ji, Xin Lin, and Chunping Liu,
\newblock ``Style transfer for anime sketches with enhanced residual u-net and
  auxiliary classifier gan,''
\newblock in {\em Asian Conference on Pattern Recognition}, 2017.

\bibitem{liu2020sketch}
Bingchen Liu, Kunpeng Song, Yizhe Zhu, and Ahmed Elgammal,
\newblock ``: Synthesizing stylized art images from sketches,''
\newblock in {\em Asian Conference on Computer Vision}, 2020.

\bibitem{scribbler}
Patsorn Sangkloy, Jingwan Lu, Chen Fang, Fisher Yu, and James Hays,
\newblock ``Scribbler: Controlling deep image synthesis with sketch and
  color,''
\newblock in {\em CVPR}, 2017.

\bibitem{edgegan}
Chengying Gao, Qi~Liu, Qi~Xu, Limin Wang, Jianzhuang Liu, and Changqing Zou,
\newblock ``Sketchycoco: Image generation from freehand scene sketches,''
\newblock in {\em CVPR}, 2020.

\bibitem{zou2019language}
Changqing Zou, Haoran Mo, Chengying Gao, Ruofei Du, and Hongbo Fu,
\newblock ``Language-based colorization of scene sketches,''
\newblock {\em ACM Transactions on Graphics (TOG)}, vol. 38, no. 6, pp. 1--16,
  2019.

\bibitem{xdog}
Holger Winnem{\"o}ller, Jan~Eric Kyprianidis, and Sven~C Olsen,
\newblock ``Xdog: an extended difference-of-gaussians compendium including
  advanced image stylization,''
\newblock {\em Computers \& Graphics}, vol. 36, no. 6, pp. 740--753, 2012.

\bibitem{hed}
Saining Xie and Zhuowen Tu,
\newblock ``Holistically-nested edge detection,''
\newblock in {\em ICCV}, 2015.

\bibitem{cyclegan}
Jun-Yan Zhu, Taesung Park, Phillip Isola, and Alexei~A. Efros,
\newblock ``Unpaired image-to-image translation using cycle-consistent
  adversarial networks,''
\newblock {\em ICCV}, 2017.

\bibitem{dualgan}
Zili Yi, Hao Zhang, Ping Tan, and Minglun Gong,
\newblock ``Dualgan: Unsupervised dual learning for image-to-image
  translation,''
\newblock {\em ICCV}, 2017.

\bibitem{pix2pix}
Phillip Isola, Jun-Yan Zhu, Tinghui Zhou, and Alexei~A. Efros,
\newblock ``Image-to-image translation with conditional adversarial networks,''
\newblock {\em CVPR}, 2017.

\bibitem{ganilla}
Samet Hicsonmez, Nermin Samet, Emre Akbas, and Pinar Duygulu,
\newblock ``Ganilla: Generative adversarial networks for image to illustration
  translation,''
\newblock {\em Image and Vision Computing}, p. 103886, 2020.

\bibitem{huang2018multimodal}
Xun Huang, Ming-Yu Liu, Serge Belongie, and Jan Kautz,
\newblock ``Multimodal unsupervised image-to-image translation,''
\newblock in {\em ECCV}, 2018.

\bibitem{coco}
Tsung-Yi Lin, Michael Maire, Serge Belongie, James Hays, Pietro Perona, Deva
  Ramanan, Piotr Doll{\'a}r, and C~Lawrence Zitnick,
\newblock ``Microsoft {COCO}: {C}ommon objects in context,''
\newblock in {\em ECCV}, 2014.

\bibitem{coco_stuff}
Holger Caesar, Jasper Uijlings, and Vittorio Ferrari,
\newblock ``Coco-stuff: Thing and stuff classes in context,''
\newblock in {\em CVPR}, 2018.

\bibitem{pytorch}
Adam Paszke, Sam Gross, Soumith Chintala, Gregory Chanan, Edward Yang, Zachary
  DeVito, Zeming Lin, Alban Desmaison, Luca Antiga, and Adam Lerer,
\newblock ``Automatic differentiation in pytorch,''
\newblock 2017.

\bibitem{adam}
Diederik~P. Kingma and Jimmy Ba,
\newblock ``Adam: A method for stochastic optimization,'' 2014.

\bibitem{ade20k}
Bolei Zhou, Hang Zhao, Xavier Puig, Sanja Fidler, Adela Barriuso, and Antonio
  Torralba,
\newblock ``Semantic understanding of scenes through the ade20k dataset,''
\newblock {\em arXiv preprint arXiv:1608.05442}, 2016.

\bibitem{FID}
Martin Heusel, Hubert Ramsauer, Thomas Unterthiner, Bernhard Nessler, and Sepp
  Hochreiter,
\newblock ``Gans trained by a two time-scale update rule converge to a local
  nash equilibrium,''
\newblock in {\em NIPS}, 2017.

\end{thebibliography}

\end{document}